\documentclass[11pt,a4paper]{article}
\usepackage[hyperref]{acl2019}
\usepackage{times}
\usepackage{latexsym}
\usepackage{graphicx}
\usepackage{hyperref}
\usepackage{amsmath}
\usepackage{amssymb}
\usepackage{multirow}
\usepackage{booktabs}
\usepackage{enumitem}
\usepackage{comment}
\usepackage{mathtools}

\usepackage{xargs}                  
\usepackage[colorinlistoftodos,prependcaption,textsize=tiny,textwidth=2.1cm]{todonotes}
\newcommandx{\jianshu}[2][1=]{\todo{#2}}

\usepackage{url}

\aclfinalcopy 

\title{Semantically Conditioned Dialog Response Generation \\ via Hierarchical Disentangled Self-Attention}

\author{Wenhu Chen$^\dagger$, Jianshu Chen$^\ddag$, Pengda Qin$^\P$, Xifeng Yan$^\dagger$ and William Yang Wang$^\dagger$ \\
$^\dagger$University of California, Santa Barbara, CA, USA\\
$^\ddag$Tencent AI Lab, Bellevue, WA, USA\\
$^\P$Beijing University of Posts and Telecommunications, China\\
{\tt \{wenhuchen,xyan,william\}@cs.ucsb.edu}\\
{\tt jianshuchen@tencent.com qinpengda@bupt.edu.cn}
}

\date{}

\begin{document}
\maketitle
\begin{abstract}
Semantically controlled neural response generation on limited-domain has achieved great performance. However, moving towards multi-domain large-scale scenarios are shown to be difficult because the possible combinations of semantic inputs grow exponentially with the number of domains. To alleviate such scalability issue, we exploit the structure of dialog acts to build a multi-layer hierarchical graph, where each act is represented as a root-to-leaf route on the graph. Then, we incorporate such graph structure prior as an inductive bias to build a hierarchical disentangled self-attention network, where we disentangle attention heads to model designated nodes on the dialog act graph. By activating different (disentangled) heads at each layer, combinatorially many dialog act semantics can be modeled to control the neural response generation. On the large-scale Multi-Domain-WOZ dataset, our model can yield a significant improvement over the baselines on various automatic and human evaluation metrics.
\end{abstract}

\section{Introduction}
\begin{figure}[t]
    \begin{center}
    \includegraphics[width=0.95\linewidth]{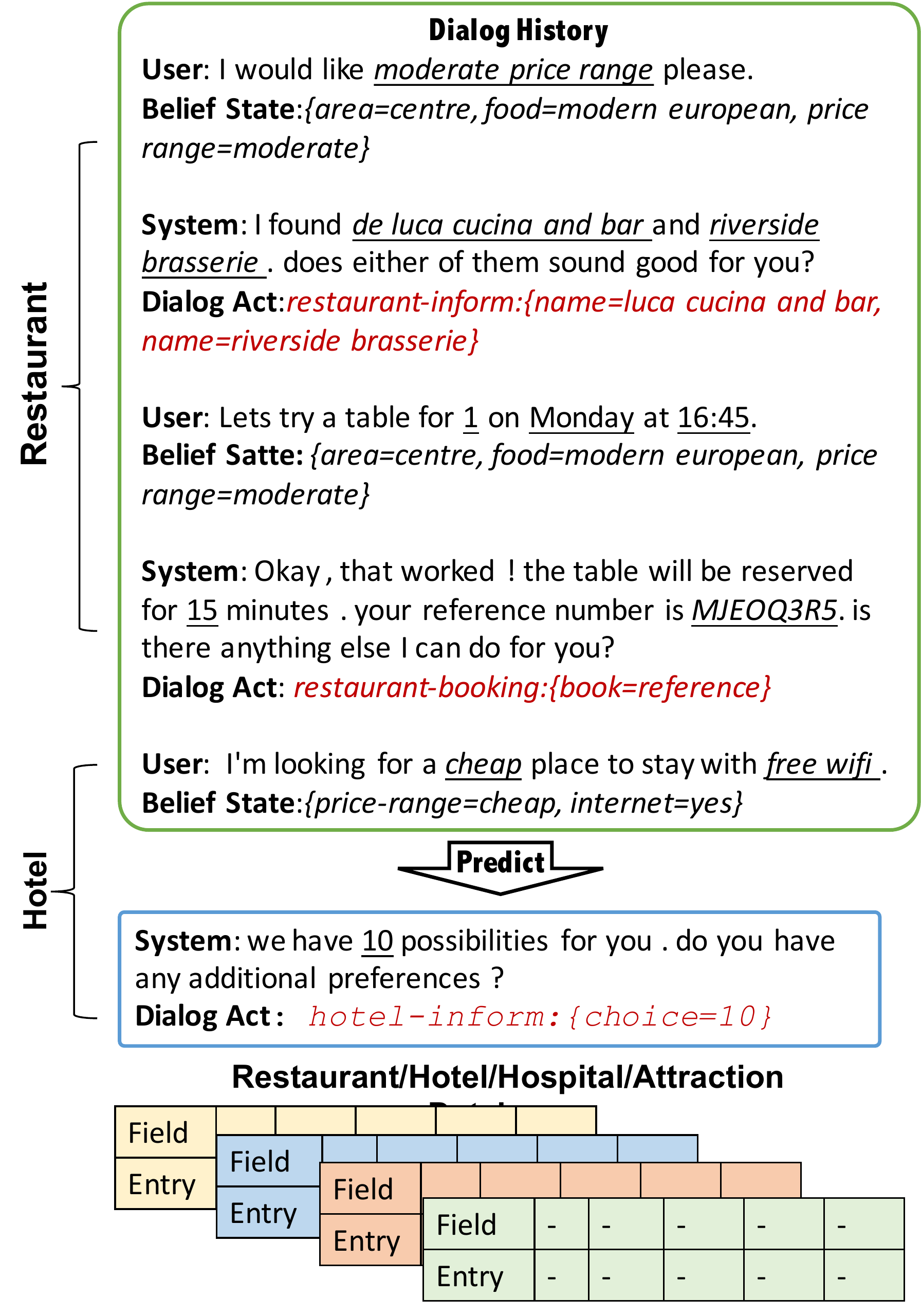}
    \end{center}
    \caption{An example dialog from MultiWOZ dataset, where the upper rectangle includes the dialog history, the tables at the bottom represent the external database, and the lower rectangle contains the dialog action and the language surface form that we need to predict.}
    \label{fig:dialog_flow}
\end{figure}

Conversational artificial intelligence~\cite{young2013pomdp} is one of the critical milestones in artificial intelligence. Recently, there have been increasing interests in industrial companies to build task-oriented conversational agents~\cite{DBLP:conf/icml/WenMBY17,DBLP:conf/ijcnlp/LiCLGC17,DBLP:conf/eacl/Rojas-BarahonaG17} to solve pre-defined tasks such as restaurant or flight bookings, etc (see~\autoref{fig:dialog_flow} for an example dialog from MultiWOZ~\cite{DBLP:conf/emnlp/BudzianowskiWTC18}). Traditional agents are built based on slot-filling techniques, which requires significant human handcraft efforts. And it is hard to generate naturally sounding utterances in a generalizable and scalable manner. Therefore, different semantically controlled neural language generation models have been developed ~\cite{DBLP:conf/emnlp/WenGMSVY15,DBLP:conf/emnlp/WenGMRSUVY16,DBLP:conf/naacl/WenGMRSVY16,DBLP:conf/sigdial/DusekJ16} to replace the traditional systems, where an explicit semantic representation (dialog act) are used to influence the RNN generation. The canonical approach is proposed in~\cite{DBLP:conf/emnlp/WenGMSVY15} to encode each individual dialog act as a unique vector and use it as an extra input feature into the cell of long short-term memory (LSTM) to influence the generation. As pointed in~\cite{DBLP:conf/naacl/WenGMRSVY16}, these models though achieving good performance on limited domains, suffer from scalability problem as the possible dialog acts grow combinatorially with the number of domains.  

\begin{figure}[t]
    \begin{center}
    \includegraphics[width=0.9\linewidth]{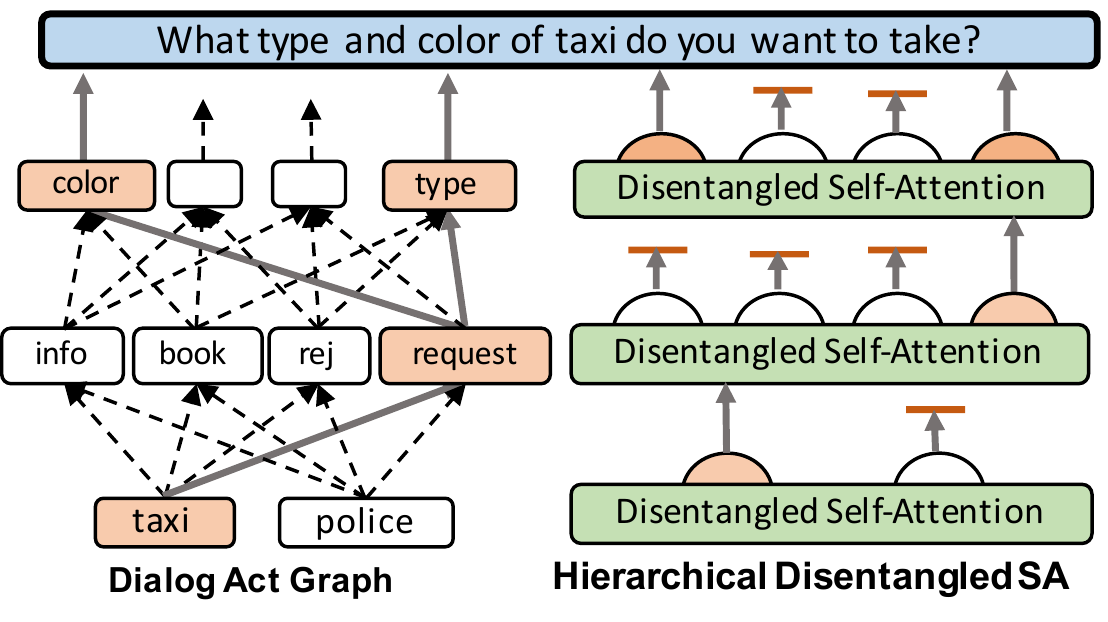}
    \end{center}
    \caption{The left part is the graph representation of the dialog acts, where each path in the graph denotes a unique dialog act. The right part denotes our proposed HDSA, where the orange nodes are activated while the others are blocked. (For details, refer to~\autoref{fig:architecture})}
    \label{fig:introduction}
\end{figure}

In order to alleviate such issue, we propose a hierarchical graph representation by leveraging the structural property of dialog acts. Specifically, we first build a multi-layer tree to represent the entire dialog act space based on their inter-relationships. Then, we merge the tree nodes with the same semantic meaning to construct an acyclic multi-layered graph, where each dialog act is interpreted as a root-to-leaf route on the graph. Such graph representation of dialog acts not only grasps the inter-relationships between different acts but also reduces the exponential representation cost to almost linear, which will also endow it with greater generalization ability. Instead of simply feeding such vectorized representation as an external feature vector to the neural networks, we propose to incorporate such a structure act as an inductive prior for designing the neural architecture, which we name as hierarchical disentangled self-attention network (HDSA).
In \autoref{fig:introduction}, we show how the dialog act graph structure is explicitly encoded into model architecture. Specifically, HDSA consists of multiple layers of disentangled self-attention modules (DSA). Each DSA has multiple switches to set the on/off state for its heads, and each head is bound for modeling a designated node in the dialog act graph. At the training stage, conditioned on the given dialog acts and the target output sentences, we only activate the heads in HDSA corresponding to the given acts (i.e., the path in the graph) to activate the heads with their designated semantics. At test time, we first predict the dialog acts and then use them to activate the corresponding heads to generate the output sequence, thereby controlling the semantics of the generated responses without handcrafting rules. As depicted in ~\autoref{fig:introduction}, by gradually activating nodes from domain $\rightarrow$ action $\rightarrow$ slot, the model is able to narrow its response down to specifically querying the user about the color and type of the taxi, which provides both strong controllability and interpretability.

Experiment results on the large-scale MultiWOZ dataset~\cite{DBLP:conf/emnlp/BudzianowskiWTC18} show that our HDSA significantly outperforms other competing algorithms.\footnote{The code and data are released in \url{https://github.com/wenhuchen/HDSA-Dialog}} In particular, the proposed hierarchical dialog act representation effectively improves the generalization ability on the unseen test cases and decreases the sample complexity on seen cases. In summary, our contributions include: (i) we propose a hierarchical graph representation of dialog acts to exploit their inter-relationships, which greatly reduces the sample complexity and improves generalization, (ii) we propose to incorporate the structure prior in semantic space to design HDSA to explicitly model the semantics of neural generation, and outperforms baselines.

\section{Related Work \& Background}
\begin{figure*}[thb]
    \begin{center}
    \includegraphics[width=1.0\linewidth]{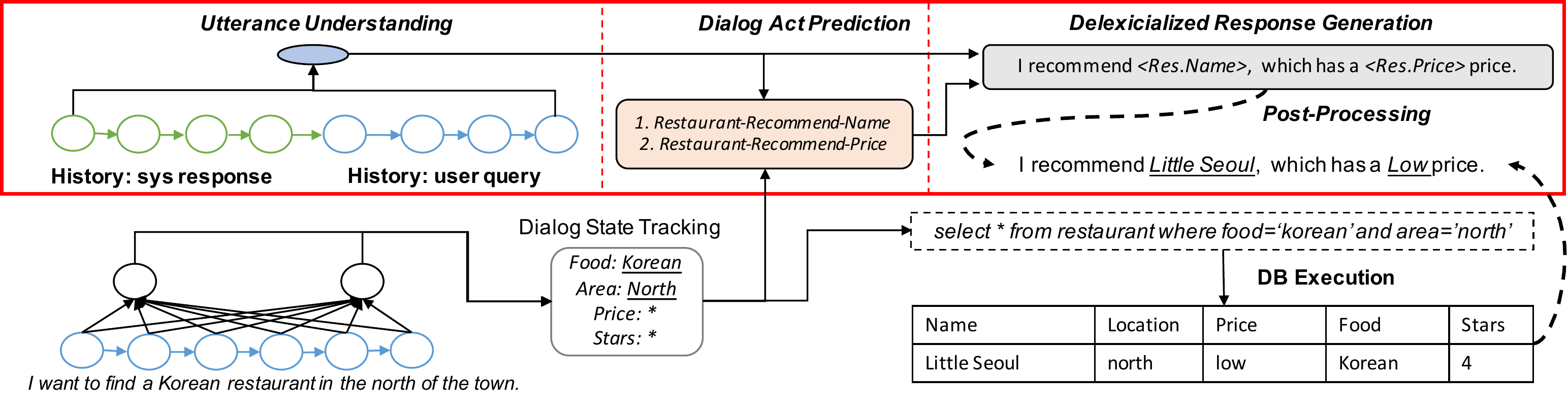}
    \end{center}
    \caption{Illustration of the neural dialog system. We decompose it into two parts: the lower part describes the dialog state tracking and DB query, and the upper part denotes the Dialog Action Prediction and Response Generation. In this paper, we are mainly interested in improving the performance of the upper part. }
    \label{fig:overview}
\end{figure*}
Canonical task-oriented dialog systems are built as pipelines of separately trained modules: (i) user intention classification~\cite{DBLP:conf/naacl/ShiYTJ16,DBLP:conf/naacl/GooGHHCHC18}, which is for understanding human intention. (ii) belief state tracker~\cite{DBLP:conf/sigdial/WilliamsRRB13,DBLP:conf/acl/MrksicSWTY17,DBLP:journals/tacl/MrksicVSLRGKY17,zhong2018global,chen2018xl}, which is used to track user's query constraint and formulate DB query to retrieve entries from a large database. (iii) dialog act prediction~\cite{DBLP:conf/icml/WenMBY17}, which is applied to classify the system action. (iv) response generation~\cite{DBLP:conf/eacl/Rojas-BarahonaG17,DBLP:conf/naacl/WenGMRSVY16,DBLP:conf/ijcnlp/LiCLGC17,DBLP:conf/acl/KanHLJRY18} to realize language surface form given the semantic constraint. In order to handle the massive number of entities in the response, ~\citet{DBLP:conf/eacl/Rojas-BarahonaG17,DBLP:conf/naacl/WenGMRSVY16,DBLP:conf/emnlp/WenGMSVY15} suggest to break response generation into two steps: first generate delexicalized sentences with placeholders like $<$Res.Name$>$, and then post-process the sentence by replacing the placeholders with the DB record. The existing modularized neural models have achieved promising performance on  limited-domain datasets like DSTC~\cite{DBLP:journals/dad/WilliamsRH16a}, CamRes767~\cite{DBLP:conf/eacl/Rojas-BarahonaG17} and KVRET~\cite{DBLP:conf/sigdial/EricKCM17}, etc. However, a recently introduced multi-domain and large-scale dataset MultiWOZ~\cite{DBLP:conf/emnlp/BudzianowskiWTC18} poses great challenges to these approaches due to the large number of slots and complex ontology. Dealing with such a large semantic space remains a challenging research problem.

We follow the nomenclature proposed in \citet{DBLP:conf/eacl/Rojas-BarahonaG17} to visualize the overview of the pipeline system in~\autoref{fig:overview}, and then decompose it into two parts: the lower part (blue rectangle) contains state tracking and symbolic DB execution, the upper part consists of dialog act prediction and response generation conditioned on the state tracking and DB results. In this paper, we are particularly interested in the upper part (act prediction and response generation) by assuming the ground truth belief state and DB records are available. More specifically, we set out to investigate how to handle the large semantic space of dialog acts and leverage it to control the neural response generation. Our approach encodes the history utterances into distributed representations to predict dialog acts and then uses the predicted dialog acts to control neural response generation. The key idea of our model is to devise a more compact structured representation of the dialog acts to reduce the exponential growth issue and then incorporate the structural prior for the semantic space into the neural architecture design. Our proposed HDSA is inspired by the linguistically-inform self-attention~\cite{DBLP:conf/emnlp/StrubellVAWM18}, which combines multi-head self-attention with multi-task NLP tasks to enhance the linguistic awareness of the model. In contrast, our model disentangles different heads to model different semantic conditions in a single task, which provides both better controllability and interpretability.
\begin{figure*}[thb]
    \begin{center}
    \includegraphics[width=1.0\linewidth]{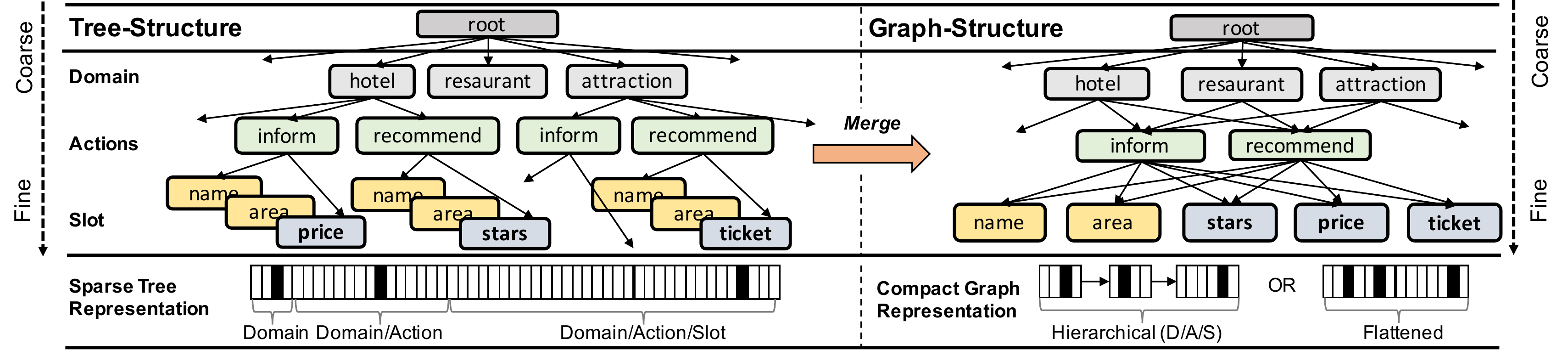}
    \end{center}
    \caption{The left figure describes the tree representation of the dialog acts, and the right figure denotes the obtained  graph representation from the left after merging the cross-branch nodes that have the same semantics. The Hierarchical form is used in our main model HDSA, Falttented is used for baseline models. }
    \label{fig:dialog-act}
    \vspace{-2ex}
\end{figure*}

\section{Dialog Act Representation}
\label{sec:act}
Dialog acts are defined as the semantic condition of the language sequence, comprising of domains, actions, slots, and values. 
\paragraph{Tree Structure}
The dialog acts
have universally hierarchical property, which is inherently due to the different semantic granularity. Each dialog act can be seen as a root-to-leaf path as depicted in~\autoref{fig:dialog-act}\footnote{we add dummy node ``none" to transform those non-leaf acts into leaf act to normalize all acts into triplet; for example ``hotel-inform" is converted into ``hotel-inform-none"}.  Such tree structure can capture the kinship between dialog acts, i.e. ``restaurant-inform-location'' has stronger similarity with ``restaurant-inform-name'' than ``hotel-request-address".  The canonical  approach to encode dialog acts is by concatenating the one-hot representation at each tree level into a flat vector like SC-LSTM~\cite{DBLP:conf/emnlp/WenGMSVY15,DBLP:conf/emnlp/BudzianowskiWTC18}  (details are in in Github\footnote{\url{https://github.com/andy194673/nlg-sclstm-multiwoz/blob/master/resource/woz3/template.txt}}). However, such representation impedes the cross-domain transfer between different slots and the cross-slot transfer between different values (e.g the ``recommend" under restaurant domain is different from ``recommend" under hospital domain). As a result, the sample complexity can grow combinatorially as the potential dialog act space expands in large-scale real-life dialog systems, where the potential domains and actions can grow dramatically. To address such issue, we propose a more compact graph representation.

\paragraph{Graph Structure}
The tree-based representation cannot capture the cross-branch relationship like ``restaurant-inform-location'' vs. ``hotel-inform-location'', leading to a huge expansion of the tree. Therefore, we propose to merge the cross-branch nodes that share the same semantics to build a compact acyclic graph in the right part of~\autoref{fig:dialog-act}\footnote{We call it graph because now one child node can have multiple parents, which violates the tree's definition.}. Formally, we let $\mathbb{A}$ denote the set of all the original dialog acts. And for each act $a\in \mathbb{A}$, we use $\mathcal{H}(a) = \{b_1, \cdots, b_i, \cdots, b_L\}$ to denote its $L$-layer graph form, where $b_i$ is its one-hot representation in the $i_{th}$ layer of the graph. For example, a dialog act ``hotel-inform-name" has a compact graph representation $\mathcal{H}(a) = \{b_1:[1,0,0], b_2:[1,0], b_3:[1,0,0,0,0]\}$. More formally, let $H_1 \cdots H_L$ denote the number of nodes at the layer of $1, \cdots, L$, respectively. Ideally, the total representation cost can be dramatically decreased from $\mathcal{O}(\prod_{i=1}^L H_i)$ tree-based representation to $H_0$=$\sum_{i=1}^L H_i$ in our graph representation. Due to the page limit, we include the full dialog act graph and its corresponding semantics in the Appendix. When multiple dialog acts $\mathcal{H}(a)_1, \cdots, \mathcal{H}(a)_k$ are involved in the single response, we propose to aggregate them as $A = BitOR(\mathcal{H}(a)_1, \cdots, \mathcal{H}(a)_k)$ as the $H_0$-dimensional graph representation, where $BitOR$ denotes the bit-wise OR operator\footnote{For example, two acts, $\mathcal{H}(a)_1=[[1,0,0],[1,0]]$ and $\mathcal{H}(a)_2=[[1,0,0],[0,1]]$, are aggregated into $A=[[1,0,0],[1,1]]$.}.

\paragraph{Generalization Ability}
Compared to the tree-based representation, the proposed graph representation under strong cross-branch overlap can greatly lower the sample complexity. Hence, it leads to great advantage under sparse training instances. For example, suppose the exact dialog act ``hotel-recommend-area" never appears in the training set. Then, at test time when used for response generation, the flat representation will obviously fail. In contrast, with our hierarchical representation, ``hotel'', ``recommend'' and ``area'' may have appeared separately in other instances (e.g., ``recommend'' appears in ``attraction-recommend-name''). Its graph representation could still be well-behaved and generalize well to the unseen (or less frequent) cases due to the strong compositionality.

\section{Model}
\begin{figure*}[thb]
    \begin{center}
    \includegraphics[width=1.0\linewidth]{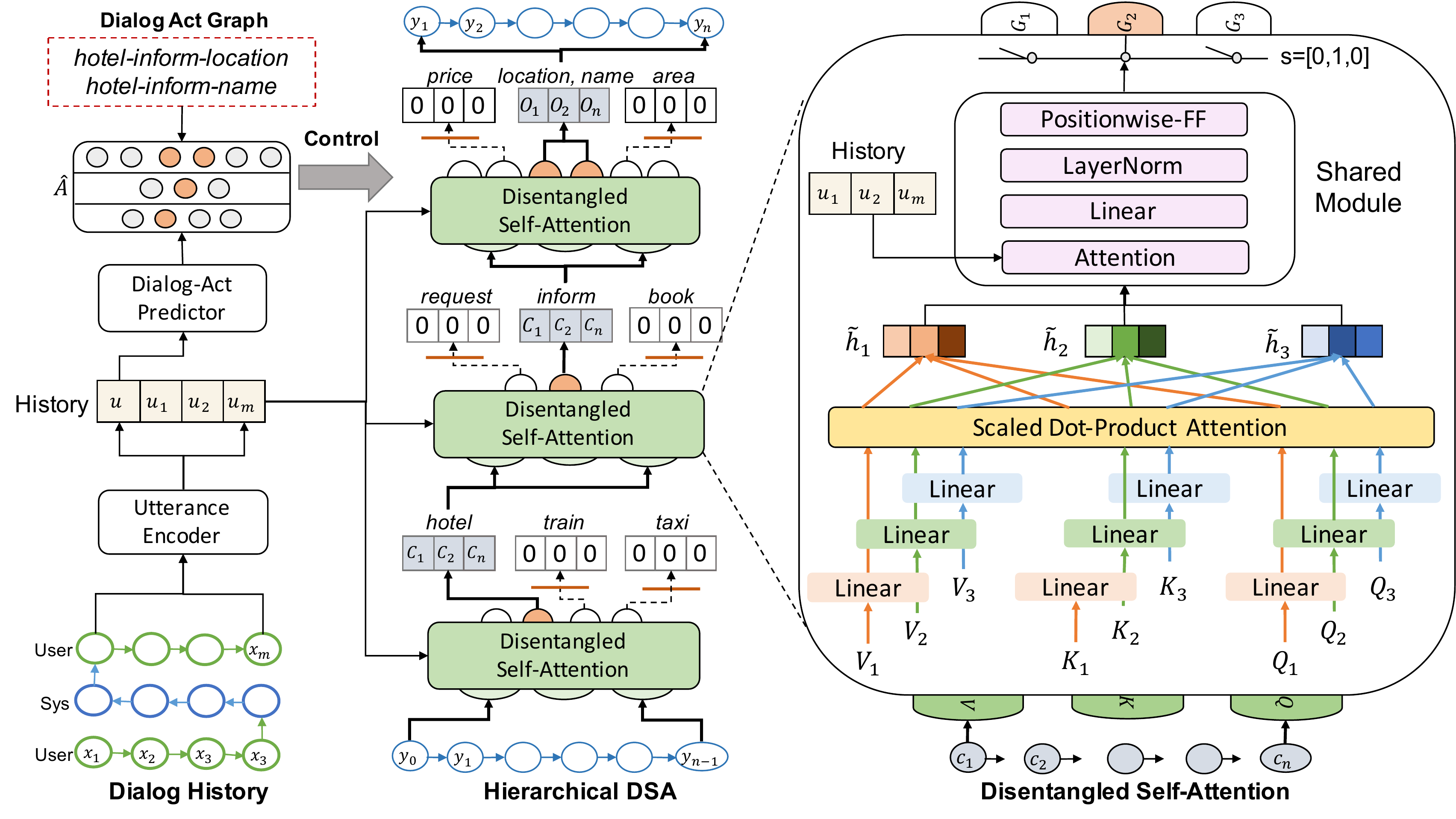}
    \end{center}
    \caption{The left figure describes the dialog act predictor and HDSA, and the right figure describes the details of DSA. The predicted hierarchical dialog acts are used to control the switch in HDSA at each layer. Here we use $L=3$ layers, the head numbers at each layer are $H=(4, 3, 6)$ heads, the hierarchical graph representation $A$=$[[0, 1, 0, 0], [0, 1, 0], [0, 0, 1, 1, 0, 0]]$. We use $m$ to denote the dialog history length and $n$ for response.}
    \label{fig:architecture}
\end{figure*}
\autoref{fig:architecture} gives an overview of our dialog system. We now proceed to discuss its components below.

\paragraph{Dialog Act Predictor}
We first explain the utterance encoder module, which uses a neural network $f_{ACT}$ to encode the dialog history (i.e., concatenation of previous utterances from both the user and the system turns $x_1, \cdots, x_m$), into distributed token-wise representations $u_1, \cdots, u_m$ with its overall representation $\bar{u}$ as follows:
\begin{align}
\small
\begin{split}
   \bar{u}, u_1, \cdots, u_m = f_{ACT}(x_1, \cdots, x_m)
\end{split}
\end{align}
where $f_{ACT}$ can be CNN, LSTM, Transformer, etc, $\bar{u}, u_1, \cdots, u_m \in \mathbb{R}^{D}$ are the representation. The overall feature $\bar{u}$ is used to predict the hierarchical representation of dialog act. That is, we output a vector $P_{\theta}(A) \in \mathbb{R}^{H_0}$, whose $i_{th}$ component gives the probability of the $i_{th}$ node in the dialog act graph being activated:
\begin{align}
\small
\begin{split}
    P_{\theta}(A) &= f_{\theta}(\bar{u}, v_{kb}, v_{bf}) \\
                  &= \sigma(V_a^T \tanh(W_u \bar{u} + W_b [v_{kb}; v_{bf}] + b))
\end{split}
\end{align}
where $V_a \in \mathbb{R}^{D \times H_0}$ is the attention matrix, the weights $W_u, W_b, b$ are the learnable parameters to project the input to $\mathbb{R}^D$ space, and $\sigma$ is the Sigmoid function. Here, we follow~\citet{DBLP:conf/emnlp/BudzianowskiWTC18,DBLP:conf/eacl/Rojas-BarahonaG17} to use one-hot vector $v_{kb}$ and $v_{bf}$ for representing the DB records and belief state (see the original papers for details). 
For convenience, we use $\theta$ to collect all the parameters of the utterance encoder and action predictor. At training time, we propose to maximize the cross-entropy objective $\mathcal{L}(\theta)$ as follows:
\begin{align}
\small
\begin{split}
    \mathcal{L}(\theta) =& A \cdot \log(f_{\theta}(\bar{u}, v_{kb}, v_{bf}) + \\
                        &(1 - A) \cdot \log(1 - f_{\theta}(\bar{u}, v_{kb}, v_{bf}))
\end{split}
\end{align}
where $\cdot$ denotes the inner product between two vectors. At test time, we predict the dialog acts $\hat{A}=\{\mathbb{I}(P_{\theta}(A)_i > T)|1\leq i\leq H_0\}$, where $T$ is the threshold and $\mathbb{I}$ is the indicator function.

\paragraph{Disentangled Self-Attention}
Recently, the self-attention-based Transformer model has achieved state-of-the-art performance on various NLP tasks such as machine translation~\cite{vaswani2017attention}, and language understanding~\cite{devlin2018bert,radford2018improving}. The success of the Transformer is partly attributed to the multi-view representation using multi-head attention architecture. Unlike the standard transformer which concatenates vectors from different heads into one vector, we propose to uses a switch to activate certain heads and only pass through their information to the next level (depicted in the right of~\autoref{fig:architecture}). Hence, we are able to disentangle the $H$ attention heads to model $H$ different semantic functionalities, and we refer to such module as the disentangled self-attention (DSA). Formally, we follow the canonical Transformer~\cite{vaswani2017attention} to define the Scaled Dot-Product Attention function given the input query/key/value features $Q, K, V \in \mathbb{R}^{n \times D}$ as: 
\begin{gather}
\small
\begin{split}
    Attention(Q, K, V) = \text{softmax}(\frac{QK^T}{\sqrt{D}})V
\end{split}
\end{gather}
where $n$ denotes the sequence length of the input, $Q, K, V$ denotes query, key and value. Here, we use $H$ different self attention functions with their independent parameterization to compute the multi-head representation $G_i$ as follows: 
\begin{gather}
\small
\begin{split}
    g_i = Attention(QW_i^Q, KW_i^K, VW_i^V)\\
    G_i = f_{PFF}(f_{LM}(f_{MLP}(f_{ATT}(g_i, u_{1:m})))
\end{split}
\end{gather}
where the input matrices $Q,K, V$ are computed from the input token embedding $x_{1:n} \in \mathbb{R}^{n \times D}$, and $D$ denotes the dimension of the embedding. The $i_{th}$ head adopts its own parameters $W_i^Q$, $W_i^K$, $W_i^V \in \mathbb{R}^{D \times \frac{D}{H}}$ to compute the output $g_i \in \mathbb{R}^{n \times  \frac{D}{H}}$. We shrink the dimension at each head to $D/H$ to reduce the computation cost as suggested in~\citet{vaswani2017attention}. 

\begin{figure}[thb]
    \begin{center}
    \includegraphics[width=0.92\linewidth]{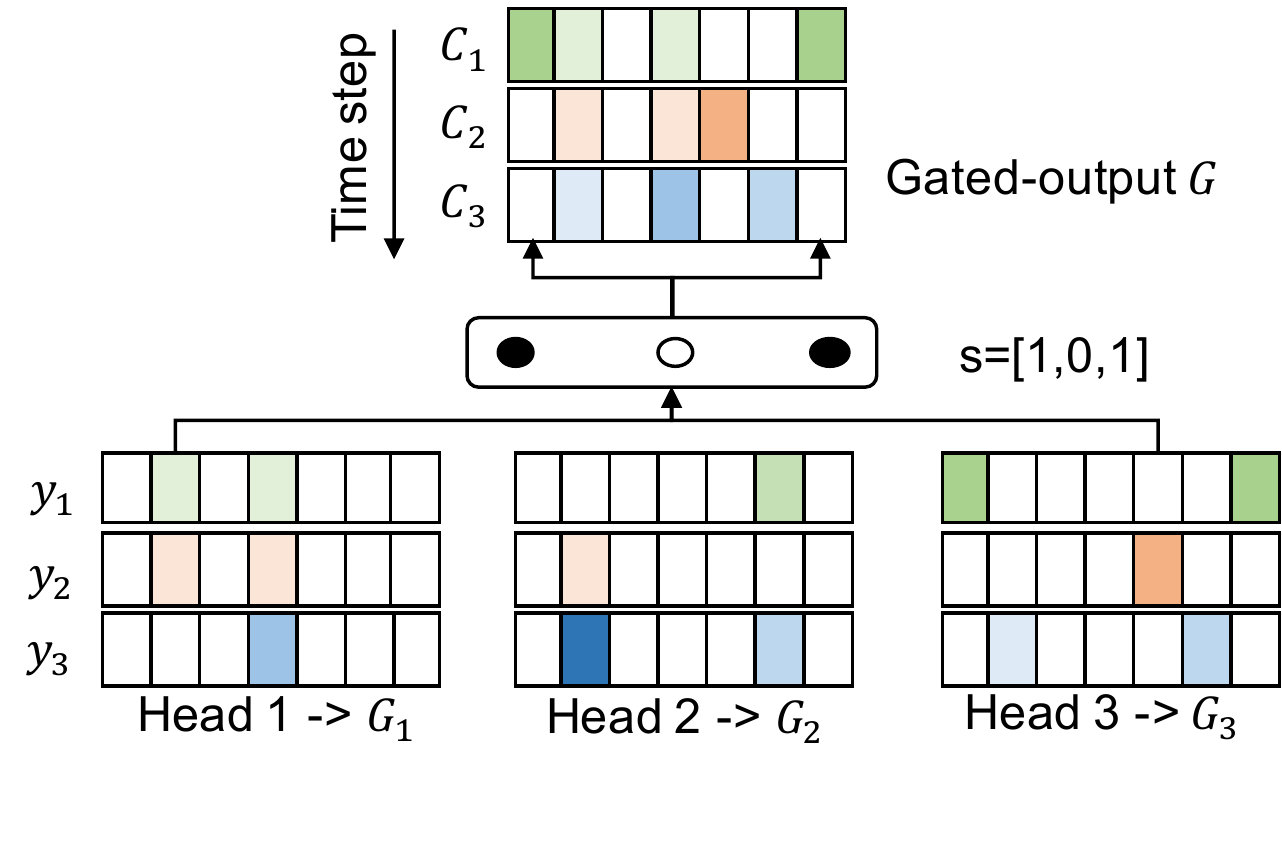}
    \end{center}
    \vspace{-4ex}
    \caption{The disentangled multi-head attention, with a sequence length of 3, 3 different heads are used with hidden dimension 7. The switch only enables the information flow from the 1st and 3rd head.}
    \label{fig:attention}
\end{figure}

We first use the cross-attention network $f_{ATT}$ to incorporate the encoded dialog history $u_{1:m}$, and then we apply a position-wise feed forward neural network $f_{PFF}$, a layer normalization $f_{LM}$, and a linear projection layer $f_{MLP}$ to obtain $G_i \in \mathbb{R}^{n \times D}$. These layers are shared across different heads. The main innovation of our architecture lies in disentangling the heads. That is, instead of concatenating $G_i$ to obtain the layer output like the standard Transformer, we employ a binary switch vector $s=(\alpha_1,\ldots,\alpha_H) \in \{0,1\}^{H}$ to control $H$ different heads and aggregate them as a $n \times D$ output matrix $G = \sum^H_{i=1} \alpha_i G_i$. Specifically, the $j$-th row of $G$, denoted as $C_j \in \mathbb{R}^D$, can be understood as the output corresponding to the $j$-th input token $y_j$ in the response. This approach is similar to a gating function to selectively pass desired information. By manipulating the attention-head switch $s$, we can better control the information flow inside the self-attention module. We illustrate the gated summation over multi-heads in~\autoref{fig:attention}.

\paragraph{Hierarchical DSA}
When the dialog system involves more complex ontology, the semantic space can grow rapidly. In consequence, a single-layer disentangled self-attention with a large number of heads is difficult to handle the complexity. Therefore, we further propose to stack multiple DSA layers to better model the huge semantic space with strong compositionality. As depicted in~\autoref{fig:overview}, the lower layers are responsible for grasping coarse-level semantics and the upper layers are responsible for capturing fine-level semantics. Such progressive generation bears a strong similarity with human brains in constructing precise responses. In each DSA layer, we feed the utterance encoding $u_{1:m}$ and last layer output $C_{1:n}$ as the input to obtain the newer output matrix $G$. We collect the output $O_{1:n}=C_{1:n}$ from the last DSA layer to compute the joint probability over a observed sequence $y_{1:n}$, which can be decomposed as a series of product over the probabilities:\footnote{We follow the standard approach in Transformer to use a mask to make $O_l$ depend only on $y_{0:l-1}$ during training. And during test time, we decode sequentially from left-to-right.}
\begin{gather*}
\small
\begin{split}
    P_{\beta}(y_{1:n}|u_{1:m}, s_{1:L}) = \prod_{l=1}^n p_{\beta}(y_l|y_{0:l-1}, u_{1:m}, s_{1:L}) \\
    p_{\beta}(y_l|y_{0:l-1}, u_{1:m}, s_{1:L}) = softmax(W_v O_l + b_v)
\end{split}
\end{gather*}
where $W_v \in \mathbb{R}^{D \times V}$ and $b_v \in \mathbb{R}^V$ are the projection weight and bias onto a vocabulary of size $V$, $l \in \{1, \cdots, n\}$ is the index, $softmax$ denotes the softmax operation, $s_{1:L}$ denotes the set of the attention switches $s_1$, $\cdots$, $s_L$ over the $L$ layers, and $\beta$ denotes all the decoder parameters. 

Recall that the graph structure of dialog acts is explicitly encoded into HDSA as a prior, where each head in HDSA is set to model a designated semantic node on the graph. In consequence, the hierarchical representation $A$ can be used to control the head switch $s_{1:L}$. At training time, the model parameters $\beta$ are optimized from the training data triple $(y_{1:n}, u_{1:m}, A)$ to maximize the likelihood of ground truth acts and responses given the dialog history. Formally, we propose to maximize the following objective function as follows:
\begin{align*}
\small
\begin{split}
    \mathcal{L}(\beta) = \log P_{\beta}(y_{1:n}|u_{1:m}, s_{1:L} = A)
\end{split}
\end{align*}
At test time, we propose to use the predicted dialog act $\hat{A}$ to control the language generation. The errors can be seen as coming from two sources, one is from inaccurate dialog act prediction, the other is from imperfect response generation. 

\section{Experiments}
\label{sec:dialog-act}
\paragraph{Dataset} 
To evaluate our proposed methods, we use the recently proposed MultiWOZ dataset~\cite{DBLP:conf/emnlp/BudzianowskiWTC18} as the benchmark, which was specifically designed to cover the challenging multi-domain and large-scale dialog managements (see the summary in~\autoref{tab:dataset}). This new benchmark involves a much larger dialog action space due to the inclusion of multiple domains and complex database backend. We represent the 625 potential dialog acts into a three-layered hierarchical graph that with a total $44$ nodes (see Appendix for the complete graph).
\begin{table}[htb]
\small
\centering
\begin{tabular}{cccc} 
\toprule
Dialogs & Total Turns   & Unique Tokens  & Value  \\
8538     & 115,424  & 24,071  &  4510     \\ 
\midrule
Dialog Acts  & Domain & Actions & Slots   \\ 
625     & 10        & 7          & 27      \\
\bottomrule
\end{tabular}
\caption{Summary of the MultiWOZ dataset.}
\label{tab:dataset}
\end{table}
We follow~\citet{DBLP:conf/emnlp/BudzianowskiWTC18} to select 1000 dialogs as the test set and 1000 dialogs as the development set. And we mainly focus on the context-to-response problem, with the dialog act prediction being a preliminary task. The best HDSA uses three DSA layers with 10/7/27 heads to separately model the semantics of domain, actions and slot (dummy head is included to model ``none" node). Adam~\cite{kingma2014adam} with a learning rate of $10^{-3}$ is used to optimize the objective. A beam size of 2 is adopted to search the hypothesis space during decoding with vocabulary size of 3,130. Also, by small-scale search, we fix the threshold $T=0.4$ due to better empirical results.
\begin{table}[htb]
\centering
\small
\begin{tabular}{llll} 
\toprule
Methods            & Precision & Recall & F1  \\ 
\midrule  
Bi-directional LSTM               &     72.4   &70.5     &  71.4 \\
Word-CNN        &       72.8    & 70.3      & 71.5    \\
3-layer Transformer &   73.3        &   72.6     &  73.1   \\
12-layer BERT &     \textbf{77.5}      &    \textbf{77.4}    &    \textbf{77.3} \\
\bottomrule
\end{tabular}
\caption{Accuracy of Dialog Act Prediction}
\label{tab:predict}
\end{table}
\begin{table*}[thb]
\small
\centering
\begin{tabular}{llcccc|c} 
\toprule
\multirow{2}{*}{Dialog-Act} & \multirow{2}{*}{Methods} & \multicolumn{4}{c|}{Delexicalized}              & Restored           \\ 
\cmidrule{3-7}
                           &                       & BLEU    & Inform & Request & \multicolumn{1}{c|}{Entity F1} & BLEU  \\ 
\midrule
\multirow{2}{*}{None}                            & LSTM~\cite{DBLP:conf/emnlp/BudzianowskiWTC18}                                 & 18.8           & 71.2 & 60.2& 54.8                           & 15.1                              \\
                                                 & 3-layer Transformer~\cite{vaswani2017attention}                         & 19.1           & 71.1 & 59.9& 55.1                           & 15.2                            \\ 
\midrule
\multirow{3}{*}{Tree Act}                         & SC-LSTM~\cite{DBLP:conf/emnlp/WenGMSVY15}              & 20.5           &74.5 & 62.5 & 57.7                           & 16.6                             \\
                                                 & 3-layer Transformer-out                               & 19.9           &74.4 & 61.1 & 57.4                           & 16.0                              \\
                                                 & 3-layer Transformer-in                                & 20.2           &73.8 & 62.1 & 57.3                           & 16.2                            \\ 
\midrule
\multirow{5}{*}{\begin{tabular}[x]{@{}l@{}}Graph Act\\(Predicted)\end{tabular}}                   & 3-layer Transformer-out                               &     22.5     & 80.8  & 64.8   &            64.2      &                      19.3   \\
                                                 & 3-layer Transformer-in                                &       22.7    &  80.4 &  65.1&                64.6           &    19.9                          \\
                                                 & Straight DSA (44 heads)  + 2 x SA          & 22.6           &  80.3  &  67.1  & 65.0                            &         20.0                     \\
                                                 & 2-layer HDSA (7/27 heads) + SA   & 23.2   & \textbf{82.9}  &  \textbf{69.1} & 65.1                  & 20.3               \\

                                                 & 3-layer HDSA (10/7/27 heads)   & \textbf{23.6}   & \textbf{82.9}  &  68.9 & \textbf{65.7}                  & \textbf{20.6}               \\
\midrule
 \multirow{3}{*}{\begin{tabular}[x]{@{}l@{}}Graph Act\\(Groundtruth)\end{tabular}}                    & 3-layer Transformer-in                          &    29.1   & 85.5 &   72.6 &            83.8   &        25.1           \\
                     & Straight DSA (44 heads)  + 2 x SA                    &  29.6     & 86.4  &  75.6 &           84.1     &            25.5    \\
                     & 3-layer HDSA (10/7/27 heads) & \textbf{30.4}   & \textbf{87.9}   &   \textbf{78.0} &        \textbf{86.2}           &    \textbf{26.2}      \\
\bottomrule
\end{tabular}
\caption{Empirical Results on MultiWOZ Response Generation, we experiment with three forms of dialog act, namely none, one-hot and hierarchical.}
\label{tab:generate}
\end{table*}%

\paragraph{Dialog Act Prediction}
We first train dialog act predictors using different neural networks to compare their performances. The experimental results (measured in F1 scores) are reported in~\autoref{tab:predict}. Experimental results show that fine-tuning the pre-trained BERT~\cite{devlin2018bert} can lead to significantly better performance than the other models. Therefore, we will use it as the dialog act prediction model in the following experiments. Instead of jointly training the predictor and the response generator, we simply fix the trained predictor when learning the generator $P_{\beta}(y)$.

\subsection{Automatic Evaluation}
We follow~\citet{DBLP:conf/emnlp/BudzianowskiWTC18} to use delexicalized-BLEU~\cite{papineni2002bleu}, inform rate and request success as three basic metrics to compare the delexicalized generation against the delexicalized reference. We further propose Entity F1~\cite{DBLP:conf/eacl/Rojas-BarahonaG17} to evaluate the entity coverage accuracy (including all slot values, days, numbers, and reference, etc), and restore-BLEU to compare the restored generation against the raw reference. 
The evaluation metrics are detailed in the supplementary material.

\begin{figure}[thb]
    \begin{center}
    \includegraphics[width=1.0\linewidth]{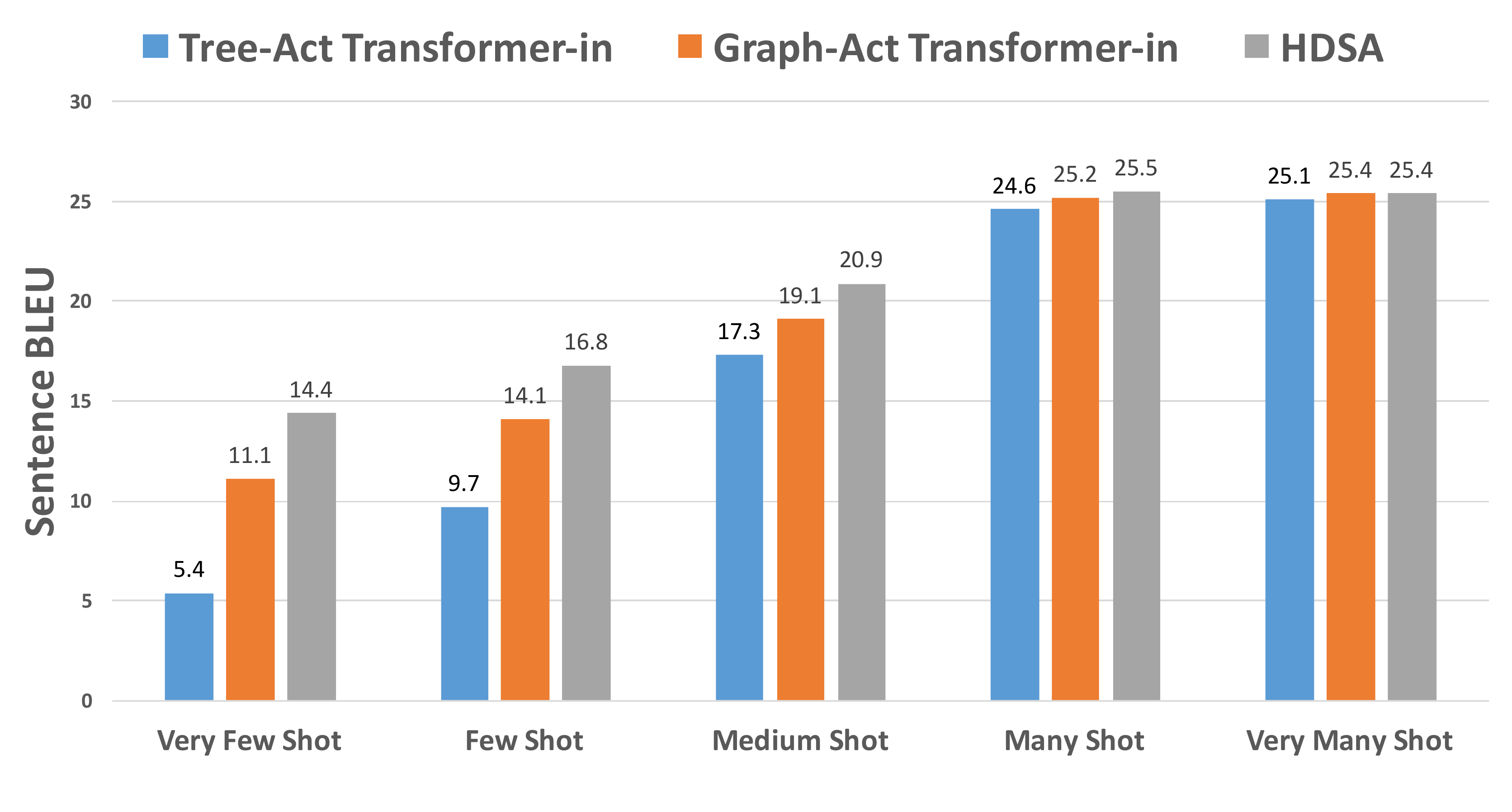}
    \end{center}
    \caption{The BLEU scores for dialog acts with different number of shots. }
    \label{fig:dialog-act-BLEU}
\end{figure}

Before diving into the experiments, we first list all the models we experiment with as follows:
\begin{enumerate}[partopsep=0pt, leftmargin=0.5cm]
    \item Without Dialog Act, we use the official code~\footnote{\url{https://github.com/budzianowski/multiwoz}}:
    (i) LSTM~\cite{DBLP:conf/emnlp/BudzianowskiWTC18}: it uses history as the attention context and applies belief state and KB results as side inputs. (ii) Transformer~\cite{vaswani2017attention}: it uses stacked Transformer architecture with dialog history as source attention context.
    \item With Sparse Tree Dialog Act, we feed the tree-based representation as an external vectors into different architectures.
    (i) SC-LSTM~\cite{DBLP:conf/emnlp/WenGMSVY15}: it feeds the sparse dialog act to the semantic gates to control the generation process.
    (ii) Transformer-in: it appends the sparse dialog act vector to input word embedding
    (iii) Transformer-out: it appends the sparse dialog act vector to the last layer output, before the softmax function.
    \item With Compact Graph Dialog Act (Predicted), we use the proposed graph representation for dialog acts and use it to control the natural language generation.
    (i) Transformer-in/out: it uses the flattened graph representation and feeds it as an external embedding feature.
    (ii) Straight DSA: it uses the flattened graph representation and model it with a one-layer DSA followed with two layers of self-attention.
    (iii) 2-layer HDSA: it adopts the partial action/slot levels of hierarchical graph representation, used as an ablation study.
    (iv) 3-layer HDSA: it adopts the full 3-layered hierarchical graph representation, used for the main model.
    \item With Graph Dialog Act (Groundtruth): it uses the ground truth dialog acts as input to see the performance upper bound of the proposed response generator architecture.
\end{enumerate}
In order to make these models comparable, we design different hidden dimensions to make their total parameter size comparable. We demonstrate the performance of different models in~\autoref{tab:generate}, and briefly conclude with the following points: (i) by feeding the sparse tree representation to input/output layer (Transformer-in/out), the model is not able to capture the large semantics space of dialog acts with sparse training instances, which unsurprisingly leads to restricted performance gain against without dialog act input. (ii) the graph dialog act is essential in reducing the sample complexity, the replacement can lead to significant and consistent improvements across different models. (iii) the hierarchical graph structure prior is an efficient inductive bias; the structure-aware HDSA can better model the compositional semantic space of dialog acts to yield a decent gain over Transformer-in/out with flattened input vector. (vi) our approaches yield significant gain (10+\%) on the Inform/Request success rate, which reflects that the explicit structured representation of dialog act is very effective in guiding dialog response in accomplishing the desired tasks. (v) the generator is greatly hindered by the predictor accuracy, by feeding the ground truth acts, the proposed HDSA is able to achieve an additional gain of 7.0 in BLEU and 21\% in Entity F1.
\paragraph{Generalization Ability}
To better understand the performance gain of the hierarchical graph-based representation, we design synthetic tests to examine its generalization ability.  Specifically, we divide the dialog acts into five categories based on their frequency of appearance in the training data: very few shot (1-100 times), few shot (100-500 times), medium shot (500-2K times), many shot (2K-5K times), and very many shot (5K+ times). We compute the average BLEU score of the turns within each frequency category and plot the result in~\autoref{fig:dialog-act-BLEU}. First, by comparing Transformer-in with compact Graph-Act against Transformer-in with sparse Tree-Act, we observe that for few number shots, the graph act significantly boosts the performance, which reflects our conjecture to lower sample complexity and generalize better to unseen (or less frequent) cases. Furthermore, by comparing Graph-Act Transformer-in with HDSA, we observe that HDSA ahieves better results by exploiting the hierarchical structure in dialog act space.
\subsection{Human Evaluation}
\paragraph{Response Quality} Owing to the low consistency between automatic metrics and human perception on conversational tasks, we also recruit trustful judges from Amazon Mechanical Turk (AMT) (with prior approval rate $>$95\%)\footnote{\url{https://www.mturk.com/}} to perform human comparison between the generated responses from HDSA and SC-LSTM. Three criteria are adopted: (i) \emph{relevance}: the response correctly answers the recent user query. (ii) \emph{coherence}: the response is coherent with the dialog history. (iii) \emph{consistency}: the generated sentence is semantically aligned with ground truth. During the evaluation, each AMT worker is presented two responses separately generated from HDSA and SC-LSTM, as well the ground truth dialog history. Each HIT assignment has 5 comparison problems, and we have a total of 200 HIT assignments to distribute. In the end, we perform statistical analysis on the harvested results after rejecting the failure cases and display the statistics in ~\autoref{tab:human}. 
\begin{table}
\small
\centering
\begin{tabular}{llll} 
\toprule
Winer       & Consistency & Relevance & Coherence  \\ 
\midrule
SC-LSTM  & 32.8\%    & 38.8\%    &36.1\%    \\
Tie         & 11.8\%    & 11.4\%    & 19.0\%     \\
HDSA     & \textbf{55.4}\%    & \textbf{49.8}\%    & \textbf{44.8}\%     \\
\bottomrule
\toprule
Model            & Match & Partial Match & Mismatch  \\ 
\midrule
HDSA &    \textbf{90\%}   &     7\%    &   \textbf{3\%}  \\
Trans-in          &  81\%    &    12\%    &   7\%  \\
SC-LSTM          &  72\%    &    10\%    &   18\%  \\
\bottomrule
\end{tabular}
\caption{Experimental results of two human evaluations for HDSA vs. SC-LSTM vs. Transformer-in. The top table gives the response quality evaluation and the bottom table demonstrates the controllability evaluation results in~\autoref{sec:controllability}.}
\label{tab:human}
\vspace{-4ex}
\end{table}
From the results, we can observe that our model significantly outperforms SC-LSTM in the \emph{coherence}, i.e., our model can better control the generation to maintain its coherence with the dialog history.
\paragraph{Semantic Controllability}\label{sec:controllability} In order to quantitatively compare the controllability of HDSA, Graph-Act Tranformer-in, and SC-LSTM, we further design a synthetic NLG experiment, where we randomly pick 50 dialog history as the context from test set, and then randomly select 3 dialog acts and their combinations as the semantic condition to control the model's responses generation. We demonstrate an example in the supplementary to visualize the evaluation procedure. Quantitatively, we hire human workers to rate (measured in match, partially match, and totally mismatch) whether the model follows the given semantic condition to generate coherent sentences. The experimental results are reported in the bottom half of~\autoref{tab:human}, which demonstrate that both the compact dialog act representation and the hierarchical structure prior are essential for controllability.

\section{Discussion}
\paragraph{Graph Representation as Transfer Learning}
The proposed graph representation works well under the cases where the set of domain slot-value pairs have significant overlaps, like Restaurant, Hotel, where the knowledge is easy to transfer. Under occasions where such exact overlap is scarce, we propose to use group similar concepts together as hypernym and use one switch to control the hypernym, which can generalize the proposed method to the broader domain. 
\paragraph{Compression vs. Expressiveness}
A trade-off that we found in our structure-based encoding scheme is that: when multiple dialog acts are involved with overlaps in the action layer, ambiguity will happen under the graph representation. For example, the two dialog acts ``restaurant-inform-price" and  ``hotel-inform-location" are merged as ``[restaurant, hotel] $\rightarrow$ [inform] $\rightarrow$ [price, location]", the current compressed representation is unable to distinguish them with ``hotel-inform-price" or ``restaurant-inform-location". Though these unnatural cases are very rare in the given dataset without hurting the performance per se, we plan to address such pending expressiveness problem in the future research.

\section{Conclusion and Future Work}
In this paper, we propose a new semantically-controlled neural generation framework to resolve the scalability and generalization problem of existing models. Currently, our proposed method only considers the supervised setting where we have annotated dialog acts, and we have not investigated the situation where such annotation is not available. In the future, we intend to infer the dialog acts from the annotated responses and use such noisy data to guide the response generation.

\section{Acknowledgements}
We really appreciate the efforts of the anonymous reviews and cherish their valuable comments, they have helped us improve the paper a lot. We are gratefully supported by a Tencent AI Lab Rhino-Bird Gift Fund. We are also very thankful for the public available dialog dataset released by University of Cambridge and PolyAI.

\bibliography{acl2019}
\bibliographystyle{acl_natbib}

\clearpage
\appendix
\section{Details of Model Implementation}
Here we detailedly explain the model implementation of the baselines and our proposed HDSA model. In the encoder side, we use a three-layered transformer with input embedding size of 64 and 4 heads, the dimension of query/value/key are all set to 16, in the output layer, the results of 4 heads are concatenated to obtain a 64-dimensional vector, which is the first broadcast into 256-dimension and then back-projected to 64-dimension. By stacking three layers of such architecture, we obtain at the end the series of 64-dimensional vectors. Following BERT, we use the first symbol as the sentence-wise representation $u$, and compute its matching score against all the tree node to predict the representation of dialog acts $\hat{A}$. 
\begin{figure}[thb]
    \begin{center}
    \includegraphics[width=1.0\linewidth]{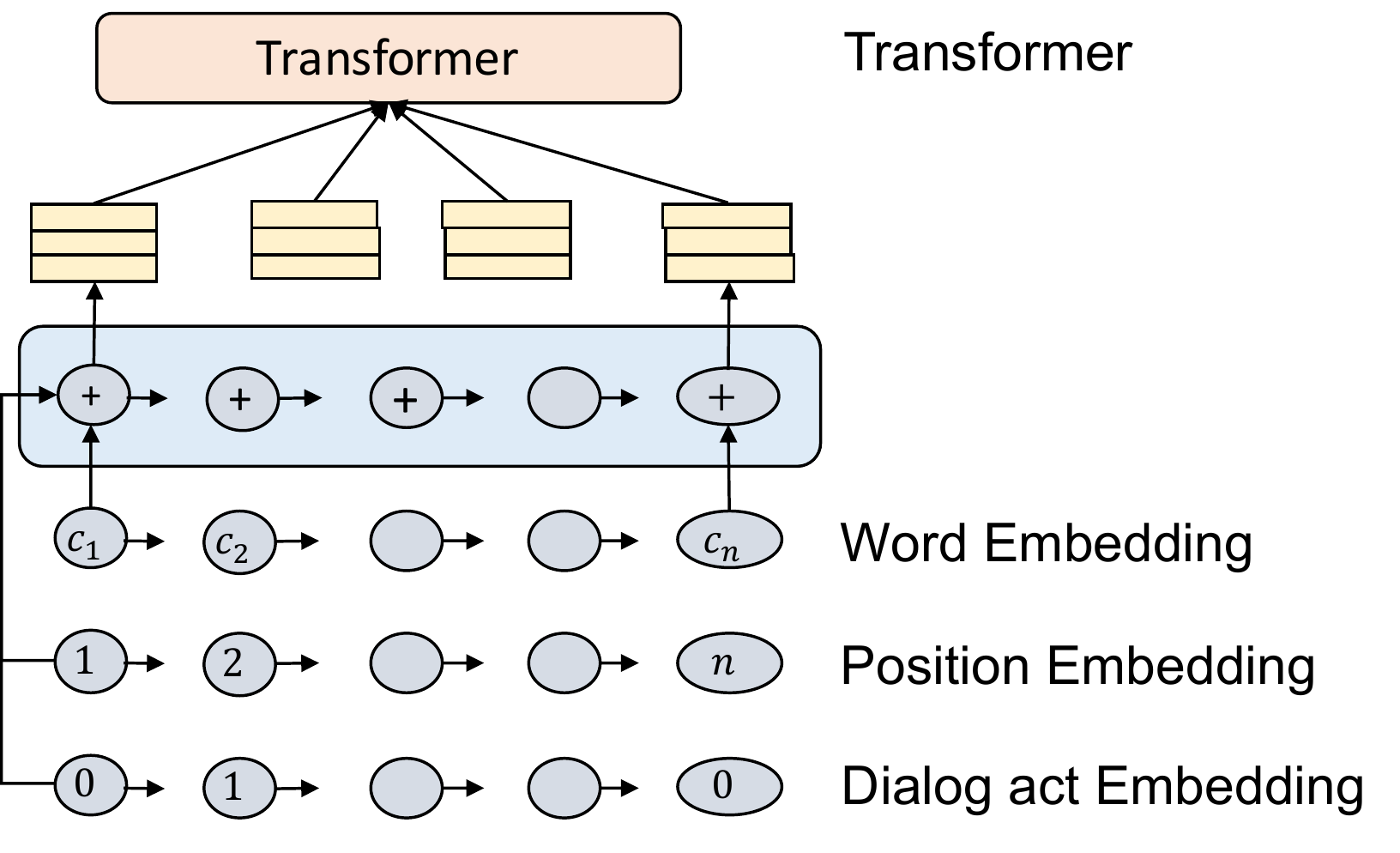}
    \end{center}
    \caption{Illustration of the architecture of Transformer-in. }
    \label{fig:transformer-in}
\end{figure}

In the decoder, we adopt take as input any length features $x_1, \cdots, x_n$, each with dimension of 64, in the first layer, since we have 10 heads, the dimension for each head is 6, thus the key, query feature dimensions are fixed to 6, the second layer with dimension of 9, the third with dimension of 2. The value feature is all fixed to 16, which is equivalent to the encoder side. After self-attention, the position-wise feed-forward neural network projects each feature back to 64 dimensions, which is further projected to 3.1K vocabulary dimension to model word probability.  

\section{Automatic Evaluation}
\begin{figure*}[thb]
    \begin{center}
    \includegraphics[width=1.0\linewidth]{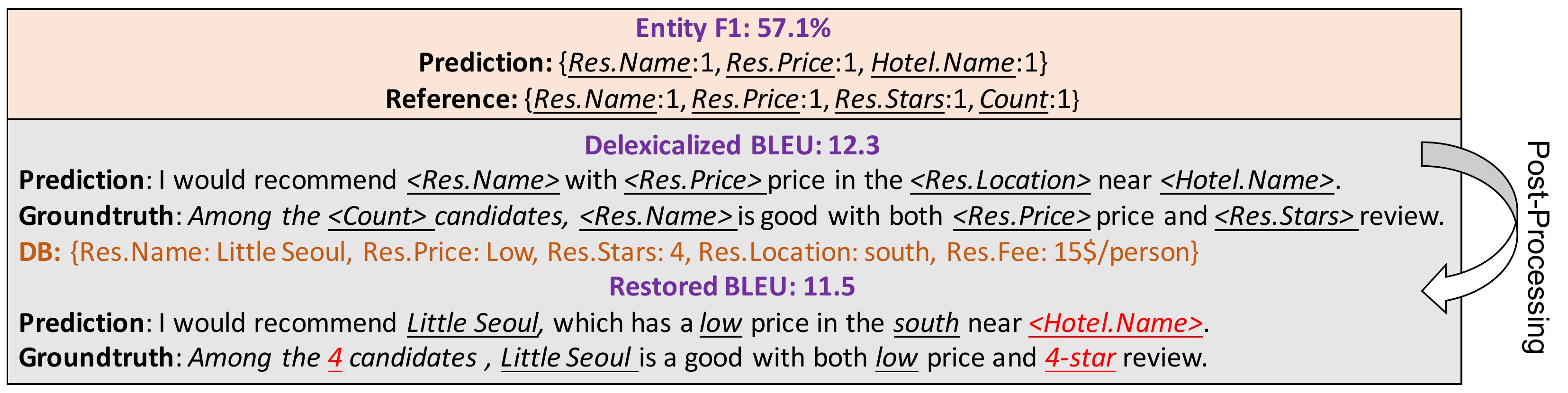}
    \end{center}
    \caption{Illustration of different evaluation metrics, in the delexicalized and non-delexicalized form.}
    \label{fig:evaluation}
\end{figure*}

We simply demonstrate an example of our automatic evaluation metrics in~\autoref{fig:evaluation}.
\label{sec:appendix}

\section{Baseline Implementation}
Here we visualize how we feed the dialog act input in as an embedding into the transformer to control the sequence generation process as~\autoref{fig:transformer-in}.

\section{Human Evaluation Interface}
\begin{figure*}[thb]
    \begin{center}
    \includegraphics[width=1.0\linewidth]{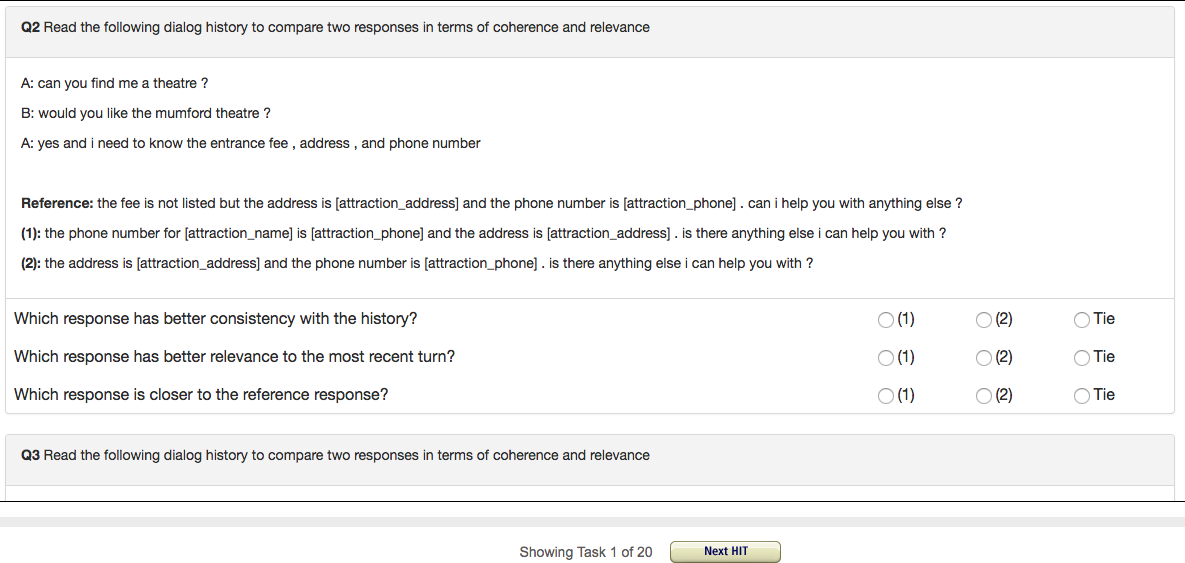}
    \end{center}
    \caption{Illustration of Human Evaluation Interface. }
    \label{fig:human-evaluate}
\end{figure*}

To better understand the human evaluation procedure, we demonstrate the user interface in~\autoref{fig:human-evaluate}.

\section{Controllability Evaluation}
\begin{figure*}[thb]
    \begin{center}
    \includegraphics[width=1.0\linewidth]{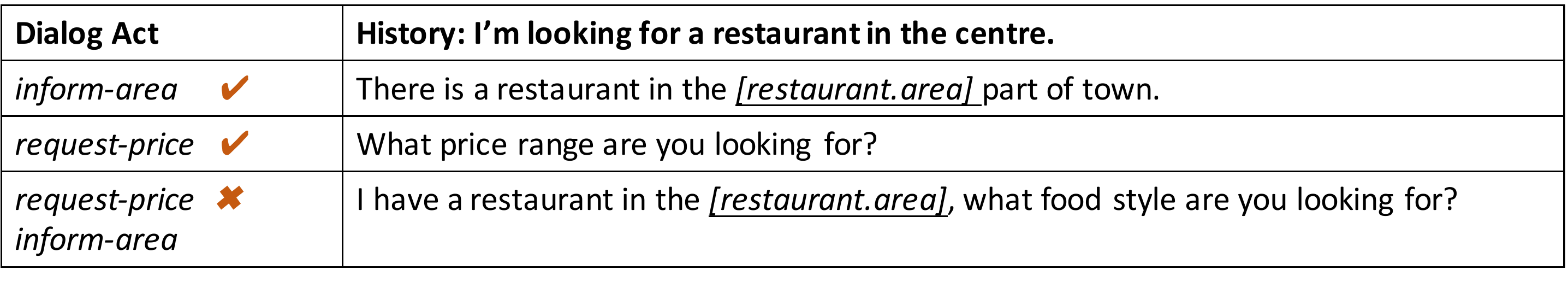}
    \end{center}
    \caption{Illustration of an example in controlling response generation given dialog act condition. Check mark means pass and cross mark means fail. }
    \label{fig:disentangled-output}
\end{figure*}

To better understand the results, we depict an example in~\autoref{fig:disentangled-output}, where 3 different dialog acts are picked as the semantic condition to constrain the response generation. 
\section{Enumeration of all the Dialog Acts}
\label{sec:enumerate}
Here we first enumerate the node semantics of the graph representation as follows:
\begin{enumerate}
    \item Domain-Layer 10 choices: \textit{'restaurant', 'hotel', 'attraction', 'train', 'taxi', 'hospital', 'police', 'bus', 'booking', 'general'.}
    \item Action-Layer 7 choices: \textit{'inform', 'request', 'recommend', 'book', 'select', 'sorry', 'none'.}
    \item Slot-Layer 27 choices: \textit{'pricerange', 'id', 'address', 'postcode', 'type', 'food', 'phone', 'name', 'area', 'choice', 'price', 'time', 'reference', 'none', 'parking', 'stars', 'internet', 'day', 'arriveby', 'departure', 'destination', 'leaveat', 'duration', 'trainid', 'people', 'department', 'stay'.}
\end{enumerate}
Then we enumerate the entire graph as follows:
\begin{figure*}[htb]
\centering
    \includegraphics[width=0.55\linewidth]{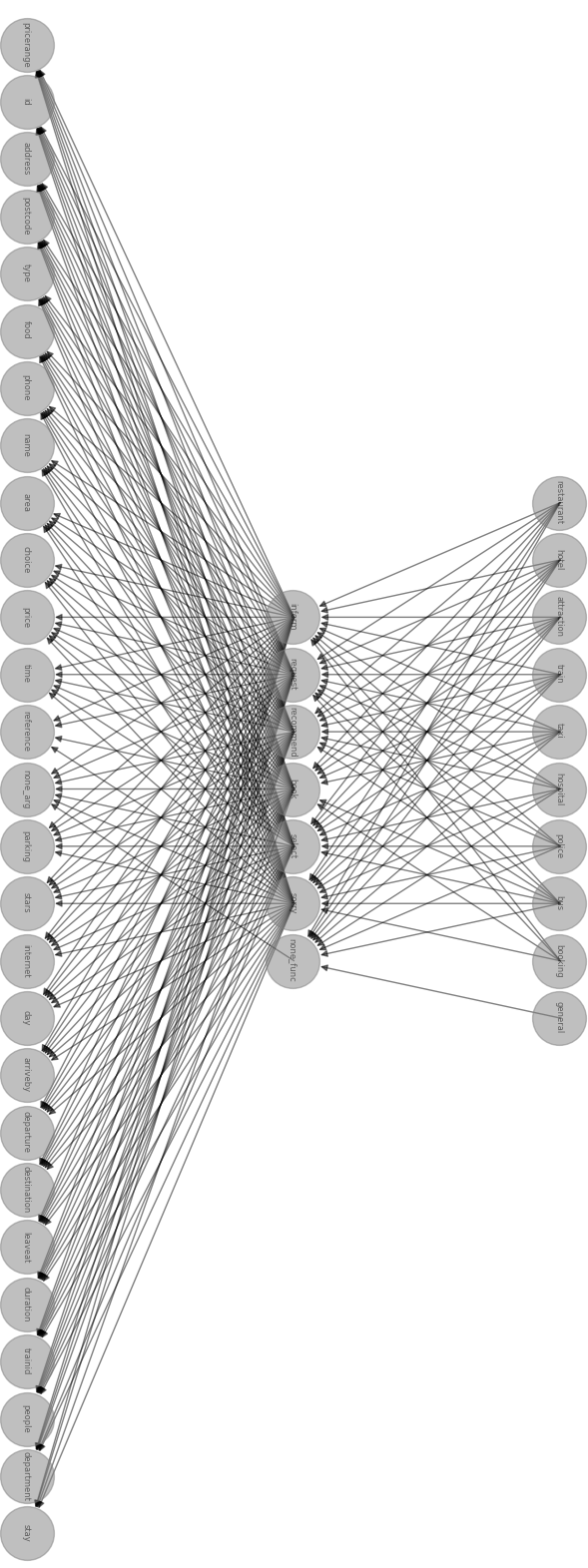}
    \caption{Illustration of entire dialog graph.}
    \label{fig:graph}
\end{figure*}

\end{document}